\documentclass[letterpaper, 10 pt, conference]{ieeeconf}  % Comment this line out if you need a4paper

\IEEEoverridecommandlockouts                              % This command is only needed if 
\usepackage{graphicx}
\usepackage{xcolor}
\usepackage{placeins}
\usepackage{todonotes}
\usepackage{booktabs} % for toprule and bottomrule in tables
\usepackage{dsfont} % Real number sign, etc.
\usepackage{tabularx} % automatic table width adjustment
\usepackage{subcaption} % subfigures
\usepackage[absolute,showboxes]{textpos}

\TPMargin*{3pt}
\newcommand\copyrighttext{
    \footnotesize
    \noindent
    SUBMITTED TO REVIEW AND POSSIBLE PUBLICATION. COPYRIGHT WILL BE TRANSFERRED WITHOUT NOTICE.\\
    Personal use of this material is permitted.
    Permission must be obtained for all other uses, in any current or future media, including reprinting/republishing this material for advertising or promotional purposes, creating new collective works, for resale or redistribution to servers or lists, or reuse of any copyrighted component of this work in other works.}%
\newcommand\copyrightnotice{%
    \begin{textblock*}{6.6in}(0.95in,0.15in)
        \centering
        \copyrighttext
    \end{textblock*}
}

\title{\LARGE \bf
Learning Through Retrospection: Improving Trajectory Prediction for Automated Driving with Error Feedback
}

\author{Steffen Hagedorn$^{1}$, Aron Distelzweig$^{2}$, Marcel Hallgarten$^{3}$, and Alexandru P. Condurache$^{1}$% <-this % stops a space
\thanks{$^{1}$Robert Bosch GmbH,
        Leonberg, Germany and
        Institute for Neuro- and Bioinformatics, University of Lübeck, Germany
        {\tt \small steffen.hagedorn@de.bosch.com}}%
\thanks{$^{2}$Department of Computer Science, University of Freiburg, Germany.}%
\thanks{$^{3}$Robert Bosch GmbH,
        Renningen, Germany and
        Cognitive Systems Group, University of Tübingen, Germany.}%
}

\begin{document}
\copyrightnotice

\maketitle
\thispagestyle{empty}
\pagestyle{empty}

%%%%%%%%%%%%%%%%%%%%%%%%%%%%%%%%%%%%%%%%%%%%%%%%%%%%%%%%%%%%%%%%%%%%%%%%%%%%%%%%
\begin{abstract}

In automated driving, predicting trajectories of surrounding vehicles supports reasoning about scene dynamics and enables safe planning for the ego vehicle.
However, existing models handle predictions as an instantaneous task of forecasting future trajectories based on observed information.
As time proceeds, the next prediction is made independently of the previous one, which means that the model cannot correct its errors during inference and will repeat them.
To alleviate this problem and better leverage temporal data, we propose a novel retrospection technique.
Through training on closed-loop rollouts the model learns to use aggregated feedback.
Given new observations it reflects on previous predictions and analyzes its errors to improve the quality of subsequent predictions.
Thus, the model can learn to correct systematic errors during inference.
Comprehensive experiments on nuScenes and Argoverse demonstrate a considerable decrease in minimum Average Displacement Error of up to $31.9\,\%$ compared to the state-of-the-art baseline without retrospection.
We further showcase the robustness of our technique by demonstrating a better handling of out-of-distribution scenarios with undetected road-users.

\end{abstract}

%%%%%%%%%%%%%%%%%%%%%%%%%%%%%%%%%%%%%%%%%%%%%%%%%%%%%%%%%%%%%%%%%%%%%%%%%%%%%%%%
\section{INTRODUCTION}
\FloatBarrier  % Ensures all figures stay within this section

\begin{figure}[tp]
    \centering
    \includegraphics[width=\columnwidth]{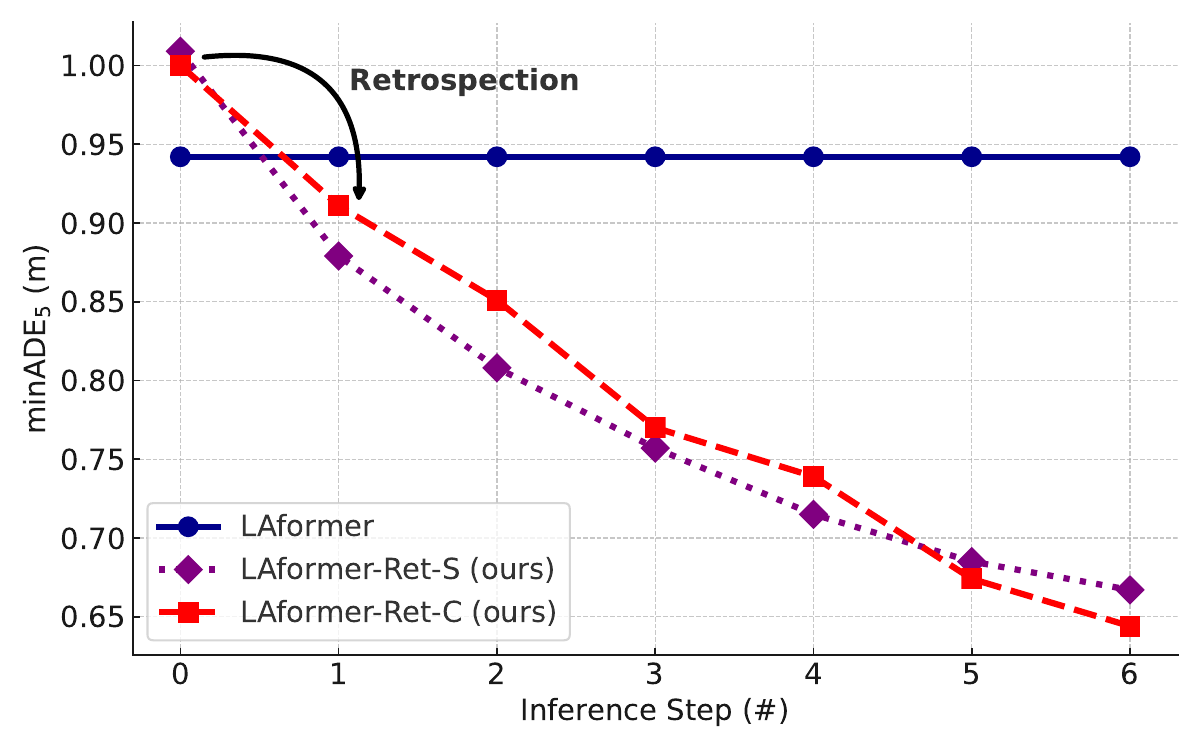}
    \caption{Learning through retrospection during inference. Feedback on how good previously made predictions match with the subsequently measured data improves the prediction quality substantially whereas conventional prediction handles consecutive predictions as independent tasks and does not improve over time. The models were trained and evaluated on nuScenes across seven consecutive predictions.}
    \label{fig:cover_figure}
\end{figure}

Accurate trajectory prediction is a cornerstone of automated driving, enabling vehicles to understand the dynamic behavior of surrounding road-users and plan safe maneuvers.
Recent advances have improved road-user interaction modeling in complex traffic environments~\cite{huang2022survey, salzmann2020trajectron++, liao2024bat}.
However, traditional prediction models treat consecutive predictions as independent tasks~\cite{lee2017desire, zhou2022hivt, liu2024laformer}.
While this approach simplifies training and inference, it ignores the potential for feedback-based learning and is therefore known as \textit{open-loop} prediction.
As a result, systematic errors are repeated in each prediction, posing risks in real-world driving scenarios.
For instance, if an observed vehicle drives more aggressively than the model learned to predict, it cannot leverage this error signal to effectively compensate for the discrepancy between the expected and observed behavior.
This limitation reveals an important insight: Integrating feedback from previous predictions can enable the system to adapt and correct its subsequent predictions.
By reflecting on discrepancies between predicted trajectories and actual measurements, a model can progressively reduce its errors, thereby aligning its predictions with the continuous, feedback-driven nature of driving.
To harness this insight, we propose a novel feedback technique that incorporates a learned retrospection module.
Through \textit{closed-loop} training, our method learns to account for its past mistakes to effectively compensate for systematic errors that might arise from model bias, missing perception information, and out-of-distribution inference.
We propose two learned retrospection modules that realize the self-correction mechanism through self-attention on the error sequence of previous predictions and cross-attention between current prediction and error sequence, respectively.
Our approach is evaluated on the challenging nuScenes~\cite{caesar2020nuscenes} and Argoverse~\cite{chang2019argoverse} datasets, demonstrating substantial reductions in minimum Average Displacement Error of up to $31.9\,\%$ compared to the baseline without retrospection (Fig.~\ref{fig:cover_figure}).
Further experiments show an improved generalization of retrospection-enhanced prediction in out-of-distribution scenarios, where input information is partially hidden from the model.
In summary, the main contributions of our work are:
\begin{enumerate}
    \item We propose a novel closed-loop training method for prediction models, enabling them to leverage error feedback from previous predictions.
    \item We introduce two model-agnostic retrospection modules to integrate feedback into subsequent predictions.
    \item We present comprehensive results on nuScenes and Argoverse, demonstrating improved prediction accuracy and robustness by continuously correcting errors during inference.
\end{enumerate}

\section{RELATED WORK}

\subsection{Trajectory Prediction} Recent advances in trajectory prediction for automated driving have centered on modeling how driving scenes evolve by leveraging rich scene representations and sophisticated interaction models~\cite{gu2021densetnt, zhou2022hivt, park2023leveraging, hallgarten2024stay, liu2024laformer, distelzweig2024entropy, hagedorn2024pioneering}. In these approaches, vectorized scene representations are processed using Graph Neural Networks~\cite{salzmann2020trajectron++, gao2020vectornet, liang2020learning, demmler2024towards} or Transformer architectures~\cite{liu2021multimodal, zhou2022hivt, ullrich2024transfer, liao2024bat}. Early methods employed one-shot prediction, i.e., generating trajectories in a single regression or classification step~\cite{cui2019multimodal, casas2018intentnet, phan2020covernet}, while more recent work has shifted toward multi-step and autoregressive formulations~\cite{liu2021multimodal, yuan2021agentformer, philion2024trajeglish} that roll out predictions sequentially.

Although autoregressive methods capture temporal dependencies more naturally, these models are still \textit{open-loop} since none of them addresses temporal dependencies spanning multiple consecutive predictions to incorporate feedback.
In contrast, we propose a \textit{closed-loop} scheme that loops an error signal back into the prediction model so that it can learn to recover from systematic errors.

Specifically, our work directly addresses this challenge by incorporating a learned feedback mechanism that uses the offsets between past predictions and subsequent observations to refine future predictions.
We term this learned ability to self-correct \textit{retrospection}.
To demonstrate the effectiveness of our model-agnostic method, we integrate it exemplarily with the state-of-the-art LAformer model~\cite{liu2024laformer}, which consists of two stages:
In its first stage, a lane-aware estimation module uses an attention mechanism at each future time step to select the top‑k candidate lane segments where the target agent is most likely to be located, ensuring that predicted motion remains aligned with the scene’s road structure. A global interaction graph then fuses vectorized representations of agent trajectories and lane segments via cross- and self-attention, and a Laplacian mixture density network decodes these fused features into multimodal trajectory predictions. In the second stage, the model refines its initial predictions by jointly considering both observed and predicted trajectories over the entire time horizon, thereby reducing errors and enhancing temporal consistency within a single prediction.

\subsection{Closed-Loop Training.} Driven by self-play and self-correction mechanisms~\cite{silver2017mastering, bardes2023v, cusumano2025robust}, closed-loop training methods have led to significant improvements in various machine learning tasks.
They involve iterative rollouts in which a model interacts with its environment and learns from the consequences of its actions. Depending on the approach, models either use supervised learning with ground-truth data or apply reinforcement learning to optimize decisions based on reward feedback.

\begin{figure}[tp]
    \centering
    \includegraphics[width=\columnwidth]{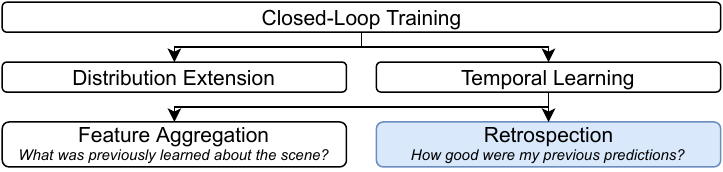}
    \caption{Advantages of closed-loop training in automated driving. Whereas many methods use closed-loop simulation to extend the distribution of training scenarios, others aggregate latent features across multiple consecutive timesteps. We propose retrospection to account for previous errors in the next prediction.}
    \label{fig:closed_loop_benefits}
\end{figure}

In automated driving, closed-loop training is implemented by simulating driving as an iterative rollout of the traffic scene~\cite{caesar2021nuplan, gulino2024waymax, yang2024drivearena}.
We identify two primary motivations for closed-loop training (Fig.~\ref{fig:closed_loop_benefits}) : extending the training data's distribution and enabling temporal learning.
Traditional \textit{open-loop} methods, such as behavior cloning~\cite{chen2015deepdriving, bojarski2016end, codevilla2019exploring}, train models to mimic recorded expert trajectories so that the training data distribution is limited to expert states.
When the model’s actions result in states that the expert never encountered, this leads to potentially catastrophic compounding errors~\cite{ross2011reduction, codevilla2018end}.
\textit{Closed-loop} training enables models to learn recovery strategies in these critical states, which is especially important in trajectory planning~\cite{zhai2023rethinking, hagedorn2024integration}.

Secondly, temporal learning aims to aggregate information across consecutive rollout steps.
Recent models, such as RealMotion~\cite{zhu2024motion}, accumulate contextual information in latent space over the course of a rollout.
This strategy allows models to remember temporarily occluded objects and refine interaction modeling~\cite{zeng2022motr, zhang2024closed}.
However, even with temporal feature aggregation, systematic errors may persist due to missing perception data or inherent model biases. To overcome these limitations, we propose a novel temporal learning technique termed \emph{retrospection}.
Unlike prior work, we do not aggregate past scene-context, but explicity loop its own previous predictions back into the model to enable reasoning about past mistakes.
By explicitly incorporating feedback on the accuracy of previous decisions—assessing “how good were my previous predictions?”—the model becomes more robust during inference.

\section{METHODOLOGY}

\begin{figure*}[tp]
    \centering
    \includegraphics[width=\textwidth]{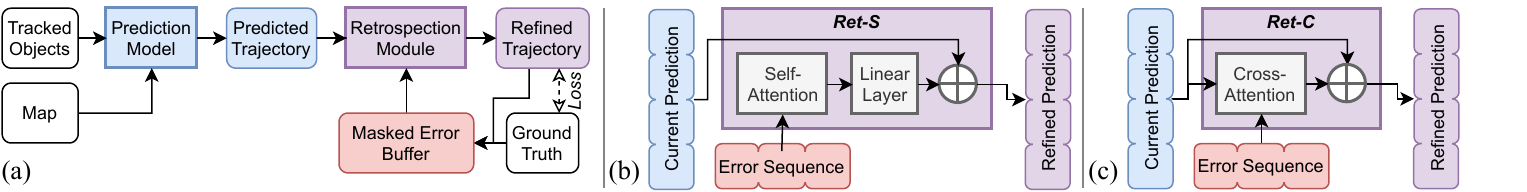}
    \caption{Learning through retrospection. (a) Model overview: the prediction model stores its previous predictions and measurements in a buffer. During training, the “measured” trajectory corresponds to the ground truth trajectory up to the current timestep (implemented as a temporal mask). A retrospection module learns to refine the trajectory by analyzing its previous predictions and how well they matched the subsequent measurements to correct systematic errors. This mechanism is then applicable at testtime. Two different retrospection modules are compared: (b) \textit{Ret-S}, which is based on self-attention of the buffered errors and (c) \textit{Ret-C}, which is based on cross-attention between the current prediction and the buffer.}
    \label{fig:system_visualization}
\end{figure*}

To investigate retrospection in practice, we extend the state-of-the-art trajectory prediction model LAformer~\cite{liu2024laformer} with a novel retrospection module and train it on closed-loop rollouts.
An overview of the whole system is depicted in Fig.~\ref{fig:system_visualization} (a).
Commonly, the task of a single-agent prediction model is defined as forecasting a set of potential future trajectories of a single target vehicle based on context information, i.e., map information and the past tracks of surrounding vehicles~\cite{janjovs2023conditional}.
In addition, we introduce a buffer that stores predicted and ground truth trajectories together with their L2 distance, i.e., the error, to implement a feedback mechanism that analyzes errors of previous predictions (Fig.~\ref{fig:error_buffer}).
Our retrospection module then uses the error signal from this buffer to correct subsequent predictions.
Since traditional methods like the Kalman filter were shown to have a limited capability to model complex dynamics~\cite{krishnan2015deep, ebert2020deep} -- as observed in traffic scenes -- we use a learned retrospection module.
Additionally, to implement closed-loop training, adaptions in data pre-processing are necessary, which are described in Sec.~\ref{sec:method-preprocessing}.

\subsection{Notation}
We define the model inputs $\mathbf{X}$ at timestep $i$ as the sequence
$\mathbf{X}^i_{-T_h:0} = \{\mathbf{X}^i_{-T_h}, \mathbf{X}^i_{-T_h+1}, \ldots, \mathbf{X}^i_{-1}, \mathbf{X}^i_{0}\}$, where $T_h$ denotes the number of past time steps.
Similarly, $\mathbf{Y}^i_{1:T_f}$ is the corresponding ground-truth, where $T_f$ denotes the number of future time steps.
% The corresponding ground truth for sample $i$ is denoted by $\mathbf{Y}^i_{1:T_f}$, where $T_f$ is the number of future time steps.
In a traditional open-loop prediction setting, a single input sample $\mathbf{X}^i_{-T_h:0}$ is provided to the model to predict a trajectory $\hat{\mathbf{Y}}^i_{1:T_f}$.
In contrast, for closed-loop training, we consider a sequence of $R$ consecutive samples, denoted $\mathbf{X}_{-T_h:0}^{1:R} = \{\mathbf{X}^1_{-T_h:0}, \mathbf{X}^2_{-T_h:0}, \ldots, \mathbf{X}^R_{-T_h:0}\}$.
Here, $R$ represents the rollout length, indicating how many consecutive samples are used for training (cf. Fig.~\ref{fig:data_preprocessing}).

\subsection{Retrospection}
\label{sec:method-retrospection}
\begin{figure}[tp]
    \centering
    \includegraphics[width=\columnwidth]{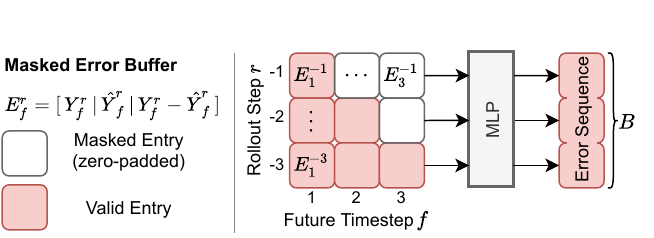}
    \caption{Masked error buffer and its conversion into an error sequence. Each buffer entry $E^r_f$ contains the concatenation of real measurement, prediction, and their pointwise distance. During inference, the masked entries are unavailable since the corresponding information was not measured yet, whereas during training the entries are zero-padded to prevent future information leakage. An MLP maps all entries of one rollout step $r$ into a single token to enable attention-based retrospection of the resulting error sequence.}
    \label{fig:error_buffer}
\end{figure}

\begin{figure}[tpb]
    \centering
    \includegraphics[width=\columnwidth]{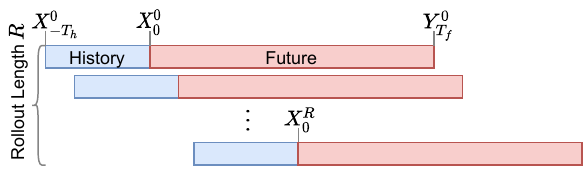}
    \caption{Data structure to enable closed-loop rollouts. We extract $R$ consecutive samples from each scenario, comprising agent track histories and ground truth future.}
    \label{fig:data_preprocessing}
\end{figure}

The core component of our work is a model-agnostic retrospection module that adapts the prediction model output based on a stored prediction error signal.
We introduce a rolling  buffer to store the error signal of the $B$ most recent rollout steps $r=-1, \dots, -B$.
Specifically, the error signal stored at rollout step $r$ is $E^r_{1:T_f}$, formed by the concatenation of model output $\hat{\mathbf{Y}}^r_{1:T_f}$, the corresponding ground truth $\mathbf{Y}^r_{1:T_f}$, and their difference.
Masking the error buffer as shown in Fig.~\ref{fig:error_buffer} avoids leakage of ground truth information during training, i.e., from the previous rollout step $r=-1$ only $\mathbf{Y}^{-1}_{1}$ is available, from the step before that $\mathbf{Y}^{-2}_{1:2}$ is available, and so on.
During live inference masking is not necessary as only measurements up to the current step are available.
To enable attention-based retrospection, all available entries $E^r_{1:T_f}$ of one rollout step $r$ are mapped into a single token by an MLP, resulting in an overall error sequence of length $B$ as depicted in Fig.~\ref{fig:error_buffer}.

In this work, we present and evaluate two different retrospection modules as shown in Fig.~\ref{fig:system_visualization}: \textbf{Ret-S} (b) and \textbf{Ret-C} (c).
\textbf{Ret-S} applies self-attention to the error buffer tokens, following the intuition that the model can analyze how its prediction error developed across rollout steps and extrapolate the expected future error from it.
After self-attention, a linear layer maps the resulting token sequence to $T_f \times 2$ offsets, which are added to the prediction model's output trajectory.

Alternatively, \textbf{Ret-C} applies cross-attention between the prediction model's output and the error buffer tokens.
Unlike in Ret-S, the new trajectory prediction is already taken into account by the attention and can be compared to previous predictions, stored in the buffer.
This way the retrospection module can adapt its offset computation to the new prediction as follows:
If it is similar to previous predictions, the error should remain similar too.
Otherwise, if the new prediction is very different from the previous one, the error will vary more.
As for Ret-S, the output sequence is projected to $T_f \times 2$ offsets with a linear layer and added to the predicted trajectory.
In both cases, a positional encoding~\cite{vaswani2017attention} is added to the inputs of the retrospection module to enable the model to understand the information in a temporal context.

\subsection{Datasets \& Metrics}
\label{sec:method-preprocessing}
We use the nuScenes~\cite{caesar2020nuscenes} and Argoverse~\cite{chang2019argoverse} datasets for our experiments to provide a broad comparison with traditional open-loop prediction models that report their results on these benchmarks.
On nuScenes, we adopt the conventional past and future horizons of $T_h=4$ and $T_f=12$ timesteps for each rollout step, corresponding to $2\,s$ and $6\,s$, respectively.
In total, $R=7$ consecutive rollout steps are extracted per scenario in the prediction split.
We discard sequences from the dataset where the target vehicle is not tracked for $R=7$ consecutive samples.
This yields $21{,}060$ scenarios out of $32{,}186$ in the training split and $5{,}982$ out of $9{,}041$ in the validation split.\\
On Argoverse, each scenario has exactly the length of a single rollout step, providing a past and future horizon of $T_h=20$ and $T_f=30$ timesteps, corresponding to $2\,s$ and $3\,s$, respectively.
To extract multiple rollout steps per scenario, we reduce the past horizon to $1.6\,s$, yielding $R=5$ steps.
Thus, all $205{,}942$ and $39{,}472$ training and validation scenarios of Argoverse can be used.
As per convention, models are trained with $K=5$ and $K=10$ modes on nuScenes, whereas $K=6$ modes are used on Argoverse.

For each dataset, we report the standard metric: minimum average displacement error (minADE), defined as the mean L2-distance between the best mode and the ground truth trajectory.
Following other works, we also report the minimum final displacement error (minFDE) on Argoverse, which is calculated like minADE but considers only the final waypoint, and miss rate (MR) on nuScenes, computed as the fraction of predictions whose best mode’s final waypoint is more than two meters away from the corresponding ground truth waypoint.

\section{RESULTS}
In Tables~\ref{tab:retrospection_nuscenes} and~\ref{tab:retrospection_argoverse} we report the results of different models and training paradigms on the nuScenes prediction split and Argoverse, respectively.
Specifically, we compare the two retrospection modules described in Sec.~\ref{sec:method-retrospection} with a baseline LAformer model trained on the same rollout data but without a retrospection module.
Additionally, we provide open-loop results for orientation as reported in other works.
For closed-loop models, results are reported at the final rollout step $R$.
We structure the discussion according to our key findings:

\textbf{1) Retrospection Iteratively Improves Predictions.}
Both retrospection modules substantially outperform the closed-loop-trained LAformer baseline in minADE -- and, to a lesser extent, in MR.
As illustrated in Fig.~\ref{fig:cover_figure}, minADE decays exponentially across inference steps: the first retrospection iteration delivers the largest reduction, with subsequent iterations yielding increasingly smaller reductions.
This pattern suggests that recent errors provide the most informative corrective signal (see insight~3).
Since only the earliest predicted waypoints can be compared to ground truth immediately -- and by the time the final waypoint error is available the trajectory is already outdated -- the model receives less informative feedback on long-term errors, explaining the smaller gain in MR (which evaluates only the final waypoint).
Nonetheless, even after a single retrospection step, both modules clearly surpass the non‑retrospective baseline, improving further to an outperformance by $31.9\, \%$ (Ret-C) in the final rollout step.

\textbf{2) Training Data Can Be Used More Efficiently.} 
On Argoverse (Table~\ref{tab:retrospection_argoverse}), we observe that the gap between the retrospection models and the baseline is smaller than on nuScenes (Table~\ref{tab:retrospection_nuscenes}).
This outcome confirms that systematic errors reduce when training on larger datasets.
Nonetheless, even with larger datasets, retrospection remains crucial for handling corner cases and improving robustness as shown in result 4).
Moreover, the closed-loop training paradigm itself -- by structuring the exact same data into more samples with a shorter history -- already outperforms open-loop training, confirming that the data structure is beneficial even in the absence of a retrospection module~\cite{li2024ego}.
This finding aligns with recent research that shows how motion forecasting tasks mainly require the current dynamic state rather than a long state history~\cite{li2024ego}.
Even when training prediction models in open-loop, the training data should consequentially be split in more samples with a shorter history.

\begin{table}[tbp]
\centering
\resizebox{1.0\columnwidth}{!}{
\begin{tabular}{l|c c|c c}
\toprule
 & \multicolumn{2}{c|}{K=5} & \multicolumn{2}{c}{K=10} \\
 & minADE & MR & minADE & MR \\ \hline
\multicolumn{5}{c}{\textit{Open-Loop Training ($32{,}186$ scenarios, R=1)}}\\
\hline
Trajectron++~\cite{salzmann2020trajectron++} & 1.88 & 0.70 & 1.51 & 0.57 \\
LaPred~\cite{kim2021lapred} & 1.47 & 0.53 & 1.12 & 0.46 \\
THOMAS~\cite{gilles2021thomas} & 1.33 & 0.55 & 1.04 & 0.42 \\
PGP~\cite{deo2022multimodal} & 1.27 & 0.52 & 0.94 & 0.34 \\
Q-EANet~\cite{chen2024q} & 1.18 & 0.48 & 1.02 & 0.44 \\
FRM~\cite{park2023leveraging} & 1.18 & 0.48 & \textbf{0.88} & 0.30 \\
Goal-LBP~\cite{yao2023goal} & \textbf{1.02} & \textbf{0.32} & 0.93 & \textbf{0.27} \\
LAformer~\cite{liu2024laformer} & 1.19 & 0.48 & 0.93 & 0.33 \\ \hline
\hline
\multicolumn{5}{c}{\textit{Open-Loop Training ($21{,}060$ scenarios, R=7)}}\\
\hline
LAformer~\cite{liu2024laformer} & 0.94 & 0.37 & 0.79 & 0.29 \\
\hline
\multicolumn{5}{c}{\textit{Closed-Loop Training ($21{,}060$ scenarios, R=7)}}\\
\hline
\textit{LAformer-Ret-S (ours)} & 0.67 & \textbf{0.35} & 0.57 & \textbf{0.25} \\
\textit{LAformer-Ret-C (ours)} & \textbf{0.64} & 0.37 & \textbf{0.55} & 0.28 \\
\bottomrule
\end{tabular}
}
\caption{Performance on nuScenes. Closed-loop results are reported at the final rollout step 7.}
\label{tab:retrospection_nuscenes}
\end{table}

\begin{table}[tbp]
\centering
\resizebox{0.74\columnwidth}{!}{
\begin{tabular}{l | c c}
\toprule
 & \multicolumn{2}{c}{K=6} \\
 & minADE & minFDE \\ \hline
\multicolumn{3}{c}{\textit{Open-Loop Training}}\\
\multicolumn{3}{c}{\textit{($205{,}942$ scenarios, $R=1$, $t_{h}=2\, s$)}}\\
\hline
TNT~\cite{zhao2021tnt} & 0.73 & 1.29 \\
LaneRCNN~\cite{zeng2021lanercnn} & 0.77 & 1.19 \\
DenseTNT~\cite{gu2021densetnt} & 0.73 & 1.05 \\
LaneGCN~\cite{liang2020learning} & 0.71 & 1.08 \\
TPCN\cite{ye2021tpcn} & 0.73 & 1.15 \\
FRM\cite{park2023leveraging} & 0.68 & 0.99 \\
HiVT~\cite{zhou2022hivt} & 0.66 & 0.96 \\
LAformer~\cite{liu2024laformer} & \textbf{0.64} & \textbf{0.92} \\ \hline
\hline
\multicolumn{3}{c}{\textit{Open-Loop Training}}\\
\multicolumn{3}{c}{\textit{($205{,}942$ scenarios, $R=5$, $t_{h}=1.6\, s$)}}\\
\hline
LAformer~\cite{liu2024laformer} & 0.58 & 1.19 \\
\hline
\multicolumn{3}{c}{\textit{Closed-Loop Training}}\\
\multicolumn{3}{c}{\textit{($205{,}942$ scenarios, $R=5$, $t_{h}=1.6\, s$)}}\\
\hline
\textit{LAformer-Ret-S (ours)} & \textbf{0.56} & \textbf{1.09} \\
\textit{LAformer-Ret-C (ours)} & \textbf{0.56} & 1.10 \\
\bottomrule
\end{tabular}
}
\caption{Performance on Argoverse. Closed-loop results are reported at final rollout step 5.}
\label{tab:retrospection_argoverse}
\end{table}

\textbf{3) Most Recent Errors are Most Informative.}
We evaluate the retrospection modules' reliance on recent errors by varying the error sequence length $B$ and training the model to convergence. A buffer of $B=6$ was used as the standard in our nuScenes experiments and retains the six most recent predictions.
As reported in Table~\ref{tab:ablation_buffer_len}, the performance gain from $B=1$ to $B=2$ exceeds that from $B=4$ to $B=6$.
This finding confirms that the most recent errors provide the most corrective information and that the information gained from further extending the retrospection horizon is limited.

\begin{table}[tbp]
\centering
\resizebox{1.0\columnwidth}{!}{
\begin{tabular}{l|c c c c}
\toprule
  & B=1 & B=2 & B=4 & B=6 \\
 \hline
\textit{LAformer-Ret-S (ours)} & 0.87 & 0.78 & 0.71 & 0.67 \\
\textit{LAformer-Ret-C (ours)} & 0.86 & 0.80 & 0.69 & \textbf{0.64} \\
\bottomrule
\end{tabular}
}
\caption{minADE$_5$ (m) on nuScenes at final rollout step 6 using different error sequence lengths B for retrospection.}
\label{tab:ablation_buffer_len}
\end{table}

\begin{figure}[tpb]
    \centering
    \includegraphics[width=\columnwidth]{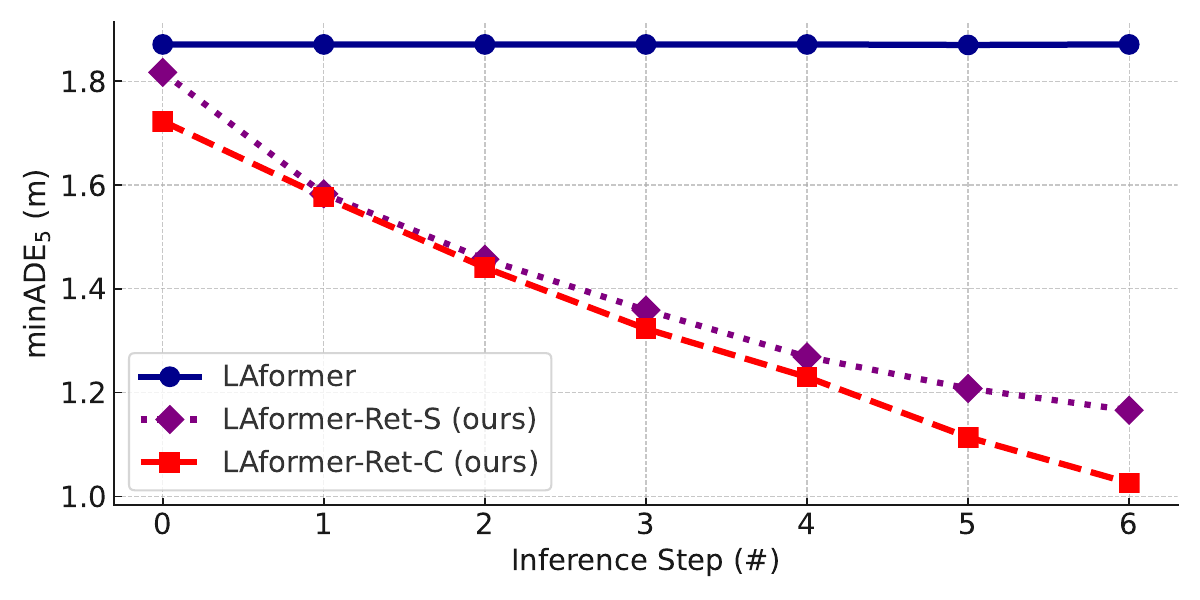}
    \caption{Inference on out-of-distribution data. We randomly drop out $10\, \%$ of the non-target agents in each nuScenes scenario at test time to simulate severe perception errors. The experiment demonstrates how retrospection increases prediction robustness: all model's initial predictions are considerably less accurate due to the missing information. While the model without retrospection performs constantly poor, the models with retrospection analyze that their predictions are inadequate and partially compensate for the influence of the "invisible" agents in the following predictions.}
    \label{fig:robustness}
\end{figure}

\textbf{4) Improved Robustness in Out-of-Distribution Scenarios.}
To assess model behavior under challenging conditions unseen during training, we simulate out-of-distribution (o.o.d.) scenarios by randomly dropping $10\,\%$ of non-target traffic agents in all rollout steps of each nuScenes validation scenario.
This setup simulates inference under severe perception errors in which not all surrounding agents are detected but still influence scene dynamics.
The results presented in Fig.~\ref{fig:robustness} show that this increases the initial minADE$_5$ for all models compared to the non-disturbed performance (Fig.~\ref{fig:cover_figure}).
Both, LAformer-Ret-S and LAformer-Ret-C, can retrospect and partially compensate for the large initial prediction errors, while the closed-loop LAformer baseline without retrospection remains at a consistently high error.\\
Compared to Fig. 1 where both models perform similarly and Ret-C only gets better for later inference steps, Ret-C outperforms Ret-S at all times in this experiment, i.e. retrospection through cross-attention is more robust.
This could be explained by its cross-attention mechanism that can analyze the dependency between the current prediction and past error signals more comprehensively.
Naturally, model output consistency decreases in o.o.d. scenarios.
Ret-S can only extrapolate error dynamics, which is harder for erratic signals.
In contrast, Ret-C has learned to model the error in dependence of the current prediction.
It can therefore compare the current prediction to previous ones and which error these caused and specifically react to different outputs.

\section{CONCLUSION AND FUTURE WORK}
We have proposed a novel closed-loop training paradigm for prediction models that enables a correction of prediction errors during inference.
The paradigm was successfully demonstrated on different datasets by adding two novel retrospection modules to the state-of-the-art prediction model LAformer.
Through a comprehensive experimental study we highlight the following benefits of our approach:
(1) Retrospection indeed corrects systematic errors, already leading to a substantial improvement of the average displacement error with only a single retrospection step available and further improvement for longer retrospection horizons.
(2)~In~out-of-distribution scenarios, where a part of the detected traffic participants is removed from the model input, retrospection helps the model to adapt to the situation and considerably improves prediction performance by learning from errors instead of repeating them.
Retrospection can consequently contribute to the safety of automated driving systems through increased robustness in novel scenes.
Combining retrospection with latent feature aggregation across timesteps is a promising future research direction to fully exploit the available information and foster the advances toward robust driving in continuous real-world conditions.

%%%%%%%%%%%%%%%%%%%%%%%%%%%%%%%%%%%%%%%%%%%%%%%%%%%%%%%%%%%%%%%%%%%%%%%%%%%%%%%%
\bibliographystyle{IEEEtran}
\bibliography{literature}

% Generated by IEEEtran.bst, version: 1.14 (2015/08/26)
\begin{thebibliography}{10}
\providecommand{\url}[1]{#1}
\csname url@samestyle\endcsname
\providecommand{\newblock}{\relax}
\providecommand{\bibinfo}[2]{#2}
\providecommand{\BIBentrySTDinterwordspacing}{\spaceskip=0pt\relax}
\providecommand{\BIBentryALTinterwordstretchfactor}{4}
\providecommand{\BIBentryALTinterwordspacing}{\spaceskip=\fontdimen2\font plus
\BIBentryALTinterwordstretchfactor\fontdimen3\font minus \fontdimen4\font\relax}
\providecommand{\BIBforeignlanguage}[2]{{%
\expandafter\ifx\csname l@#1\endcsname\relax
\typeout{** WARNING: IEEEtran.bst: No hyphenation pattern has been}%
\typeout{** loaded for the language `#1'. Using the pattern for}%
\typeout{** the default language instead.}%
\else
\language=\csname l@#1\endcsname
\fi
#2}}
\providecommand{\BIBdecl}{\relax}
\BIBdecl

\bibitem{huang2022survey}
Y.~Huang, J.~Du, Z.~Yang, Z.~Zhou, L.~Zhang, and H.~Chen, ``A survey on trajectory-prediction methods for autonomous driving,'' \emph{IEEE Transactions on Intelligent Vehicles}, vol.~7, no.~3, pp. 652--674, 2022.

\bibitem{salzmann2020trajectron++}
T.~Salzmann, B.~Ivanovic, P.~Chakravarty, and M.~Pavone, ``Trajectron++: Dynamically-feasible trajectory forecasting with heterogeneous data,'' in \emph{Computer Vision--ECCV 2020: 16th European Conference, Glasgow, UK, August 23--28, 2020, Proceedings, Part XVIII 16}.\hskip 1em plus 0.5em minus 0.4em\relax Springer, 2020, pp. 683--700.

\bibitem{liao2024bat}
H.~Liao, Z.~Li, H.~Shen, W.~Zeng, D.~Liao, G.~Li, and C.~Xu, ``Bat: Behavior-aware human-like trajectory prediction for autonomous driving,'' in \emph{Proceedings of the AAAI Conference on Artificial Intelligence}, vol.~38, no.~9, 2024, pp. 10\,332--10\,340.

\bibitem{lee2017desire}
N.~Lee, W.~Choi, P.~Vernaza, C.~B. Choy, P.~H. Torr, and M.~Chandraker, ``Desire: Distant future prediction in dynamic scenes with interacting agents,'' in \emph{Proceedings of the IEEE conference on computer vision and pattern recognition}, 2017, pp. 336--345.

\bibitem{zhou2022hivt}
Z.~Zhou, L.~Ye, J.~Wang, K.~Wu, and K.~Lu, ``Hivt: Hierarchical vector transformer for multi-agent motion prediction,'' in \emph{Proceedings of the IEEE/CVF Conference on Computer Vision and Pattern Recognition}, 2022, pp. 8823--8833.

\bibitem{liu2024laformer}
M.~Liu, H.~Cheng, L.~Chen, H.~Broszio, J.~Li, R.~Zhao, M.~Sester, and M.~Y. Yang, ``Laformer: Trajectory prediction for autonomous driving with lane-aware scene constraints,'' in \emph{Proceedings of the IEEE/CVF Conference on Computer Vision and Pattern Recognition}, 2024, pp. 2039--2049.

\bibitem{caesar2020nuscenes}
H.~Caesar, V.~Bankiti, A.~H. Lang, S.~Vora, V.~E. Liong, Q.~Xu, A.~Krishnan, Y.~Pan, G.~Baldan, and O.~Beijbom, ``nuscenes: A multimodal dataset for autonomous driving,'' in \emph{Proc. of IEEE/CVF CVPR}, 2020, pp. 11\,621--11\,631.

\bibitem{chang2019argoverse}
M.-F. Chang, J.~Lambert, P.~Sangkloy, J.~Singh, S.~Bak, A.~Hartnett, D.~Wang, P.~Carr, S.~Lucey, D.~Ramanan \emph{et~al.}, ``Argoverse: 3d tracking and forecasting with rich maps,'' in \emph{Proceedings of the IEEE/CVF conference on computer vision and pattern recognition}, 2019, pp. 8748--8757.

\bibitem{gu2021densetnt}
J.~Gu, C.~Sun, and H.~Zhao, ``Densetnt: End-to-end trajectory prediction from dense goal sets,'' in \emph{Proceedings of the IEEE/CVF International Conference on Computer Vision}, 2021, pp. 15\,303--15\,312.

\bibitem{park2023leveraging}
D.~Park, H.~Ryu, Y.~Yang, J.~Cho, J.~Kim, and K.-J. Yoon, ``Leveraging future relationship reasoning for vehicle trajectory prediction,'' \emph{arXiv:2305.14715}, 2023.

\bibitem{hallgarten2024stay}
M.~Hallgarten, I.~Kisa, M.~Stoll, and A.~Zell, ``Stay on track: A frenet wrapper to overcome off-road trajectories in vehicle motion prediction,'' in \emph{2024 IEEE Intelligent Vehicles Symposium (IV)}.\hskip 1em plus 0.5em minus 0.4em\relax IEEE, 2024, pp. 795--802.

\bibitem{distelzweig2024entropy}
A.~Distelzweig, A.~Look, E.~Kosman, F.~Janjo{\v{s}}, J.~Wagner, and A.~Valada, ``Entropy-based uncertainty modeling for trajectory prediction in autonomous driving,'' \emph{arXiv:2410.01628}, 2024.

\bibitem{hagedorn2024pioneering}
S.~Hagedorn, M.~Milich, and A.~P. Condurache, ``Pioneering se (2)-equivariant trajectory planning for automated driving,'' in \emph{2024 IEEE Intelligent Vehicles Symposium (IV)}.\hskip 1em plus 0.5em minus 0.4em\relax IEEE, 2024, pp. 2097--2102.

\bibitem{gao2020vectornet}
J.~Gao, C.~Sun, H.~Zhao, Y.~Shen, D.~Anguelov, C.~Li, and C.~Schmid, ``Vectornet: Encoding hd maps and agent dynamics from vectorized representation,'' in \emph{Proceedings of the IEEE/CVF conference on computer vision and pattern recognition}, 2020, pp. 11\,525--11\,533.

\bibitem{liang2020learning}
M.~Liang, B.~Yang, R.~Hu, Y.~Chen, R.~Liao, S.~Feng, and R.~Urtasun, ``Learning lane graph representations for motion forecasting,'' in \emph{Computer Vision--ECCV 2020: 16th European Conference, Glasgow, UK, August 23--28, 2020, Proceedings, Part II 16}.\hskip 1em plus 0.5em minus 0.4em\relax Springer, 2020, pp. 541--556.

\bibitem{demmler2024towards}
T.~Demmler, A.~Tamke, T.~Dang, K.~Haug, and L.~Mikelsons, ``Towards consistent and explainable motion prediction using heterogeneous graph attention,'' in \emph{2024 IEEE Intelligent Vehicles Symposium (IV)}.\hskip 1em plus 0.5em minus 0.4em\relax IEEE, 2024, pp. 168--175.

\bibitem{liu2021multimodal}
Y.~Liu, J.~Zhang, L.~Fang, Q.~Jiang, and B.~Zhou, ``Multimodal motion prediction with stacked transformers,'' in \emph{Proceedings of the IEEE/CVF conference on computer vision and pattern recognition}, 2021, pp. 7577--7586.

\bibitem{ullrich2024transfer}
L.~Ullrich, A.~McMaster, and K.~Graichen, ``Transfer learning study of motion transformer-based trajectory predictions,'' in \emph{2024 35th IEEE Intelligent Vehicles Symposium (IV)}.\hskip 1em plus 0.5em minus 0.4em\relax Jeju Island, Korea: IEEE, 2024, pp. 110--117.

\bibitem{cui2019multimodal}
H.~Cui, V.~Radosavljevic, F.-C. Chou, T.-H. Lin, T.~Nguyen, T.-K. Huang, J.~Schneider, and N.~Djuric, ``Multimodal trajectory predictions for autonomous driving using deep convolutional networks,'' in \emph{2019 ICRA}.\hskip 1em plus 0.5em minus 0.4em\relax IEEE, 2019, pp. 2090--2096.

\bibitem{casas2018intentnet}
S.~Casas, W.~Luo, and R.~Urtasun, ``Intentnet: Learning to predict intention from raw sensor data,'' in \emph{Conference on Robot Learning}.\hskip 1em plus 0.5em minus 0.4em\relax PMLR, 2018, pp. 947--956.

\bibitem{phan2020covernet}
T.~Phan-Minh, E.~C. Grigore, F.~A. Boulton, O.~Beijbom, and E.~M. Wolff, ``Covernet: Multimodal behavior prediction using trajectory sets,'' in \emph{Proceedings of the IEEE/CVF conference on computer vision and pattern recognition}, 2020, pp. 14\,074--14\,083.

\bibitem{yuan2021agentformer}
Y.~Yuan, X.~Weng, Y.~Ou, and K.~M. Kitani, ``Agentformer: Agent-aware transformers for socio-temporal multi-agent forecasting,'' in \emph{Proceedings of the IEEE/CVF international conference on computer vision}, 2021, pp. 9813--9823.

\bibitem{philion2024trajeglish}
J.~Philion, X.~B. Peng, and S.~Fidler, ``Trajeglish: Traffic modeling as next-token prediction,'' in \emph{The Twelfth International Conference on Learning Representations}, 2024.

\bibitem{silver2017mastering}
D.~Silver, J.~Schrittwieser, K.~Simonyan, I.~Antonoglou, A.~Huang, A.~Guez, T.~Hubert, L.~Baker, M.~Lai, A.~Bolton \emph{et~al.}, ``Mastering the game of go without human knowledge,'' \emph{nature}, vol. 550, no. 7676, pp. 354--359, 2017.

\bibitem{bardes2023v}
A.~Bardes, Q.~Garrido, J.~Ponce, X.~Chen, M.~Rabbat, Y.~LeCun, M.~Assran, and N.~Ballas, ``V-jepa: Latent video prediction for visual representation learning,'' 2023.

\bibitem{cusumano2025robust}
M.~Cusumano-Towner, D.~Hafner, A.~Hertzberg, B.~Huval, A.~Petrenko, E.~Vinitsky, E.~Wijmans, T.~Killian, S.~Bowers, O.~Sener \emph{et~al.}, ``Robust autonomy emerges from self-play,'' \emph{arXiv:2502.03349}, 2025.

\bibitem{caesar2021nuplan}
H.~Caesar, J.~Kabzan, K.~S. Tan, W.~K. Fong, E.~Wolff, A.~Lang, L.~Fletcher, O.~Beijbom, and S.~Omari, ``nuplan: A closed-loop ml-based planning benchmark for autonomous vehicles,'' \emph{arXiv:2106.11810}, 2021.

\bibitem{gulino2024waymax}
C.~Gulino, J.~Fu, W.~Luo, G.~Tucker, E.~Bronstein, Y.~Lu, J.~Harb, X.~Pan, Y.~Wang, X.~Chen \emph{et~al.}, ``Waymax: An accelerated, data-driven simulator for large-scale autonomous driving research,'' \emph{Advances in Neural Information Processing Systems}, vol.~36, 2024.

\bibitem{yang2024drivearena}
X.~Yang, L.~Wen, Y.~Ma, J.~Mei, X.~Li, T.~Wei, W.~Lei, D.~Fu, P.~Cai, M.~Dou \emph{et~al.}, ``Drivearena: A closed-loop generative simulation platform for autonomous driving,'' \emph{arXiv:2408.00415}, 2024.

\bibitem{chen2015deepdriving}
C.~Chen, A.~Seff, A.~Kornhauser, and J.~Xiao, ``Deepdriving: Learning affordance for direct perception in autonomous driving,'' in \emph{Proc. of IEEE ICCV}, 2015, pp. 2722--2730.

\bibitem{bojarski2016end}
M.~Bojarski, ``End to end learning for self-driving cars,'' \emph{arXiv:1604.07316}, 2016.

\bibitem{codevilla2019exploring}
F.~Codevilla, E.~Santana, A.~M. L{\'o}pez, and A.~Gaidon, ``Exploring the limitations of behavior cloning for autonomous driving,'' in \emph{Proceedings of the IEEE/CVF international conference on computer vision}, 2019, pp. 9329--9338.

\bibitem{ross2011reduction}
S.~Ross, G.~Gordon, and D.~Bagnell, ``A reduction of imitation learning and structured prediction to no-regret online learning,'' in \emph{Proceedings of the fourteenth international conference on artificial intelligence and statistics}.\hskip 1em plus 0.5em minus 0.4em\relax JMLR Workshop and Conference Proceedings, 2011, pp. 627--635.

\bibitem{codevilla2018end}
F.~Codevilla, M.~M{\"u}ller, A.~L{\'o}pez, V.~Koltun, and A.~Dosovitskiy, ``End-to-end driving via conditional imitation learning,'' in \emph{2018 IEEE international conference on robotics and automation (ICRA)}.\hskip 1em plus 0.5em minus 0.4em\relax IEEE, 2018, pp. 4693--4700.

\bibitem{zhai2023rethinking}
J.-T. Zhai, Z.~Feng, J.~Du, Y.~Mao, J.-J. Liu, Z.~Tan, Y.~Zhang, X.~Ye, and J.~Wang, ``Rethinking the open-loop evaluation of end-to-end autonomous driving in nuscenes,'' \emph{arXiv:2305.10430}, 2023.

\bibitem{hagedorn2024integration}
S.~Hagedorn, M.~Hallgarten, M.~Stoll, and A.~P. Condurache, ``The integration of prediction and planning in deep learning automated driving systems: A review,'' \emph{IEEE Transactions on Intelligent Vehicles}, 2024.

\bibitem{zhu2024motion}
X.~Zhu, ``Motion forecasting in continuous driving,'' in \emph{The Thirty-eighth Annual Conference on Neural Information Processing Systems}, 2024.

\bibitem{zeng2022motr}
F.~Zeng, B.~Dong, Y.~Zhang, T.~Wang, X.~Zhang, and Y.~Wei, ``Motr: End-to-end multiple-object tracking with transformer,'' in \emph{European Conference on Computer Vision}.\hskip 1em plus 0.5em minus 0.4em\relax Springer, 2022, pp. 659--675.

\bibitem{zhang2024closed}
Z.~Zhang, P.~Karkus, M.~Igl, W.~Ding, Y.~Chen, B.~Ivanovic, and M.~Pavone, ``Closed-loop supervised fine-tuning of tokenized traffic models,'' \emph{arXiv:2412.05334}, 2024.

\bibitem{janjovs2023conditional}
F.~Janjo{\v{s}}, M.~Hallgarten, A.~Knittel, M.~Dolgov, A.~Zell, and J.~M. Z{\"o}llner, ``Conditional unscented autoencoders for trajectory prediction,'' \emph{arXiv:2310.19944}, 2023.

\bibitem{krishnan2015deep}
R.~G. Krishnan, U.~Shalit, and D.~Sontag, ``Deep kalman filters,'' \emph{arXiv:1511.05121}, 2015.

\bibitem{ebert2020deep}
J.~Ebert, T.~Gumpp, S.~M{\"u}nzner, A.~Matskevych, A.~P. Condurache, and C.~Gl{\"a}ser, ``Deep radar sensor models for accurate and robust object tracking,'' in \emph{2020 IEEE 23rd International Conference on Intelligent Transportation Systems (ITSC)}.\hskip 1em plus 0.5em minus 0.4em\relax IEEE, 2020, pp. 1--6.

\bibitem{vaswani2017attention}
A.~Vaswani, ``Attention is all you need,'' \emph{Advances in Neural Information Processing Systems}, 2017.

\bibitem{li2024ego}
Z.~Li, Z.~Yu, S.~Lan, J.~Li, J.~Kautz, T.~Lu, and J.~M. Alvarez, ``Is ego status all you need for open-loop end-to-end autonomous driving?'' in \emph{Proceedings of the IEEE/CVF Conference on Computer Vision and Pattern Recognition}, 2024, pp. 14\,864--14\,873.

\bibitem{kim2021lapred}
B.~Kim, S.~H. Park, S.~Lee, E.~Khoshimjonov, D.~Kum, J.~Kim, J.~S. Kim, and J.~W. Choi, ``Lapred: Lane-aware prediction of multi-modal future trajectories of dynamic agents,'' in \emph{Proceedings of the IEEE/CVF Conference on Computer Vision and Pattern Recognition}, 2021, pp. 14\,636--14\,645.

\bibitem{gilles2021thomas}
T.~Gilles, S.~Sabatini, D.~Tsishkou, B.~Stanciulescu, and F.~Moutarde, ``Thomas: Trajectory heatmap output with learned multi-agent sampling,'' \emph{arXiv:2110.06607}, 2021.

\bibitem{deo2022multimodal}
N.~Deo, E.~Wolff, and O.~Beijbom, ``Multimodal trajectory prediction conditioned on lane-graph traversals,'' in \emph{Conference on Robot Learning}.\hskip 1em plus 0.5em minus 0.4em\relax PMLR, 2022, pp. 203--212.

\bibitem{chen2024q}
J.~Chen, Z.~Wang, J.~Wang, and B.~Cai, ``Q-eanet: Implicit social modeling for trajectory prediction via experience-anchored queries,'' \emph{IET Intelligent Transport Systems}, vol.~18, no.~6, pp. 1004--1015, 2024.

\bibitem{yao2023goal}
Z.~Yao, X.~Li, B.~Lang, and M.~C. Chuah, ``Goal-lbp: Goal-based local behavior guided trajectory prediction for autonomous driving,'' \emph{IEEE Transactions on Intelligent Transportation Systems}, 2023.

\bibitem{zhao2021tnt}
H.~Zhao, J.~Gao, T.~Lan, C.~Sun, B.~Sapp, B.~Varadarajan, Y.~Shen, Y.~Shen, Y.~Chai, C.~Schmid \emph{et~al.}, ``Tnt: Target-driven trajectory prediction,'' in \emph{Conference on Robot Learning}.\hskip 1em plus 0.5em minus 0.4em\relax PMLR, 2021, pp. 895--904.

\bibitem{ye2021tpcn}
M.~Ye, T.~Cao, and Q.~Chen, ``Tpcn: Temporal point cloud networks for motion forecasting,'' in \emph{Proceedings of the IEEE/CVF Conference on Computer Vision and Pattern Recognition}, 2021, pp. 11\,318--11\,327.

\end{thebibliography}

\end{document}